\documentclass[10pt,twocolumn,letterpaper]{article}

\usepackage{iccv}
\usepackage{times}
\usepackage{epsfig}
\usepackage{graphicx}
\usepackage{amsmath}
\usepackage{amssymb}
\usepackage{array}
\usepackage{tabu}
\usepackage{multirow}
\usepackage{color,xcolor}
\usepackage{caption} 

\usepackage[pagebackref=true,breaklinks=true,letterpaper=true,colorlinks,bookmarks=false]{hyperref}

\iccvfinalcopy % *** Uncomment this line for the final submission

\def\iccvPaperID{3624} % *** Enter the ICCV Paper ID here

\ificcvfinal\fi

\begin{document}

%%%%%%%%% TITLE
\title{Feature-Fused Context-Encoding Network for Neuroanatomy Segmentation}

\author{Yuemeng Li, Hangfan Liu, Hongming Li, and Yong Fan\\
University of Pennsylvania\\
{\tt\small ymli@seas.upenn.edu, \{hangfan.liu, hongming.li, yong.fan\}@uphs.upenn.edu }
% \and
}

\maketitle
%\thispagestyle{empty}

%need to modify the results into 88.8 for fused

%%%%%%%%% ABSTRACT
\begin{abstract}
Automatic segmentation of fine-grained brain structures remains a challenging task. Current segmentation methods mainly utilize 2D and 3D deep neural networks. The 2D networks take image slices as input to produce coarse segmentation in less processing time, whereas the 3D networks take the whole image volumes to generated fine-detailed segmentation with more computational burden. In order to obtain accurate fine-grained segmentation efficiently, in this paper, we propose an end-to-end Feature-Fused Context-Encoding Network for brain structure segmentation from MR (magnetic resonance) images. Our model is implemented based on a 2D convolutional backbone, which integrates a 2D encoding module to acquire planar image features and a spatial encoding module to extract spatial context information. A global context encoding module is further introduced to capture global context semantics from the fused 2D encoding and spatial features. The proposed network aims to fully leverage the global anatomical prior knowledge learned from context semantics, which is represented by a structure-aware attention factor to recalibrate the outputs of the network. In this way, the network is guaranteed to be aware of the class-dependent feature maps to facilitate the segmentation. We evaluate our model on 2012 Brain Multi-Atlas Labelling Challenge dataset for 134 fine-grained structure segmentation. Besides, we validate our network on 27 coarse structure segmentation tasks. Experimental results have demonstrated that our model can achieve improved performance compared with the state-of-the-art approaches.

\end{abstract}

%-------------------------------------------------------------------------
\begin{figure}[!h]
\begin{center}
\includegraphics[width=0.48\textwidth]{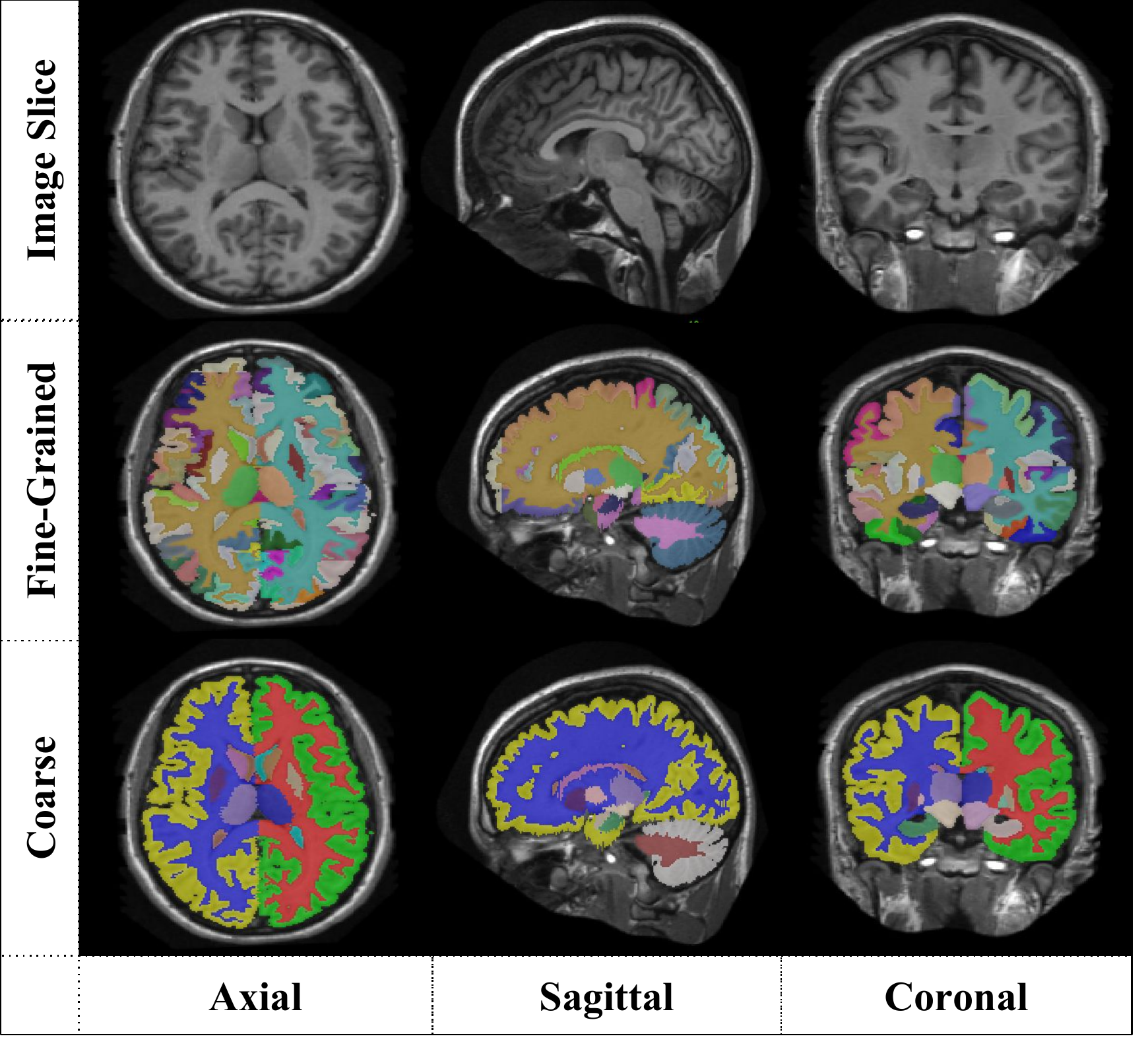}
\end{center}
  \caption{The image slice, fine-grained and coarse structure segmentation labels along three principle axes.}
\label{view}
\end{figure}
%-------------------------------------------------------------------------

%%%%%%%%% BODY TEXT
\section{Introduction}
%talk about MRI
Brain structure segmentation from magnetic resonance (MR) images provides a powerful tool for characterizing in-vivo neuroanatomy in an objective and reproducible manner, which is important for neuroimaging studies regarding brain development, aging, and brain diseases \cite{stoub2005mri,ostby2009heterogeneity}. Though several solutions have been proposed for automatic brain structure segmentation, it is still very challenging to obtain precise segmentation due to overlapped image intensity between brain structures, partial volume effects in MR images, and inter-subject anatomical variability.

With highly competitive accuracy, atlas based segmentation methods \cite{hao2014local,wang2013multi} have been widely used in medical image segmentation tasks. Instead of performing pixel/voxel-wise prediction for brain structure labels, they leverage the spatial and anatomical context information from the atlas images and labels, and therefore are more robust to intensity variations and noises within images of different subjects. However, as atlas based segmentation requires multiple deformable registration between the image to be segmented and the atlas images, it is time consuming and computationally expensive to obtain segmentation results for a large cohort of subjects. 

Along the huge success of deep learning techniques in image analysis tasks, such as classification, semantic segmentation \etc \cite{krizhevsky2012imagenet,he2017mask,long2015fully,chen2014semantic,zheng2015conditional,lafferty2001conditional}, deep learning approaches have attracted great attention in medical image analysis research. Deep learning based approaches for semantic and medical image segmentation are mostly based on the fully convolutional networks (FCN) architecture. The informative feature representation from local image regions can be captured hierarchically from the convolutional and pooling layers, and the up-sample layer is utilized to restore high-level semantic features at finer spatial scale for dense prediction. However, the rich contextual information from images is limited to local context only, even though the feature representation learned within FCN has already enclosed the intrinsic context information. While several techniques are designed to enlarge the receptive field by multi-resolution pyramid-based strategy \cite{zhao2017pyramid,chen2018deeplab,chen2018encoder}, the contextual information could not be obtained by simply increasing receptive fields.

Recent state-of-the-art methods have exploited different network structures based on multi-view 2D image slices \cite{zhang2015deep,nie2016fully,akkus2017deep,roy2019recalibrating,roy2019quicknat,roy2017error}. Although such methods have achieved promising performance, they discard all the intrinsic 3D context information when learning the feature representations for structure semantics, while it is obvious that 3D data can provide more informative context information. Moreover, existing brain structure segmentation models usually adopt relatively coarse structure labels, \eg less than 30 labels \cite{wachinger2018deepnat,roy2019quicknat,roy2019recalibrating}, even when fine-grained labels (more than 100 structures) are available. Whereas such fine-grained segmentation could provide more detailed measurements regarding neuroanatomy, it is more challenging as half of these structures are shown in small volumes with similar image appearances.

In order to facilitate the fine-grained segmentation, we propose a novel end-to-end Feature-Fused Context-Encoding network to fully leverage the global anatomical prior and utilize the spatial context information. Particularly, a modified global context encoding module \cite{zhang2018context} is incorporated to handle anatomical prior globally. A structure-aware attention factor is learned to recalibrate the output of the decoder path with little extra computational burden. Furthermore, parallel to the feature representation learned from the single input image slice, consecutive image slices are used as additional 2D multi-channel input for an auxiliary encoder path to encode spatial context information. This strategy takes advantage of 2D convolutional layers to extract the 3D spatial context for computational efficiency without loss of the spatial details. 

We evaluate the proposed model on the 2012 Multi-Atlas Labelling Challenge (MALC) dataset. Empirical results demonstrate that the proposed scheme achieves state-of-the-art performance on both coarse and fine-grained brain structure segmentation.

In summary, the contributions of this paper include:

\begin{enumerate}
    \item We propose the very first end-to-end Feature-Fused Context-Encoding network that leverages context encoding for MRI segmentation. Our network utilizes a single view 2D image slice along with the 3D consecutive spatial context. A semantic encoding classification loss is incorporated to enforce the learning of global information.
    \item We find the optimal backbone structure and class weights for the network.
    \item Our proposed model achieves a new state-of-the-art performance on MALC dataset in both coarse (27 labels) and fine-grained (134 labels) brain structure segmentation tasks.
    \item Our model segments a 3D MRI Brain scan in 6 seconds, which is 3 times faster than the recent state-of-the-art method.
\end{enumerate}

%-------------------------------------------------------------------------
\section{Related Works}

\subsection{Semantic Segmentation}
Inspired by FCN, several elegantly designed frameworks have been proposed to advance the performance of semantic segmentation \cite{chen2018deeplab,zhao2017pyramid,chen2018encoder,zhang2018context,zheng2015conditional,lafferty2001conditional}. Different from natural image segmentation, MRI brain structure segmentation is performed on 3D volumetric dataset, and each voxel is represented by a gray-scale intensity value. Previous works on brain structure segmentation have favored volumetric segmentation based on 3D inputs \cite{de2015deep,moeskops2016automatic,wachinger2018deepnat,kamnitsas2017efficient,brosch2016deep,li2017compactness}. One of the state-of-the-art works DeepNAT \cite{wachinger2018deepnat} focuses on predicting labels from 3D inputs under a hierarchical classification and multi-task learning setting. DeepNAT was operated on overlapped 3D image patches, which largely hampers its computational efficiency. 

Recent literature utilize 2D images as inputs to achieve fast processing speed without loss of accuracy. QuickNAT \cite{roy2019quicknat} and the recalibrated QuickNAT \cite{roy2019recalibrating} utilize the modified U-Net framework \cite{ronneberger2015u} with densely connected blocks \cite{huang2017densely} operating on multi-view (Coronal, Axial, Sagittal) 2D image slices. Experimental results have demonstrated QuickNAT could obtain the whole brain segmentation in about 20 seconds while achieving state-of-the-art accuracy. Their studies show that the Dense U-Net structure is especially effective on gray-scale 3D volumetric segmentation. The densely connected blocks are used to aid gradient flow and promote feature re-usability, which is crucial when the number of training data is limited. Inspired by the success of this structure, we use the QuickNAT as the baseline model, and implement the same dense block structure inside each encoder and decoder for our proposed network.

%-------------------------------------------------------------------------
\begin{figure*}[!t]
\begin{center}
\includegraphics[width=1\textwidth]{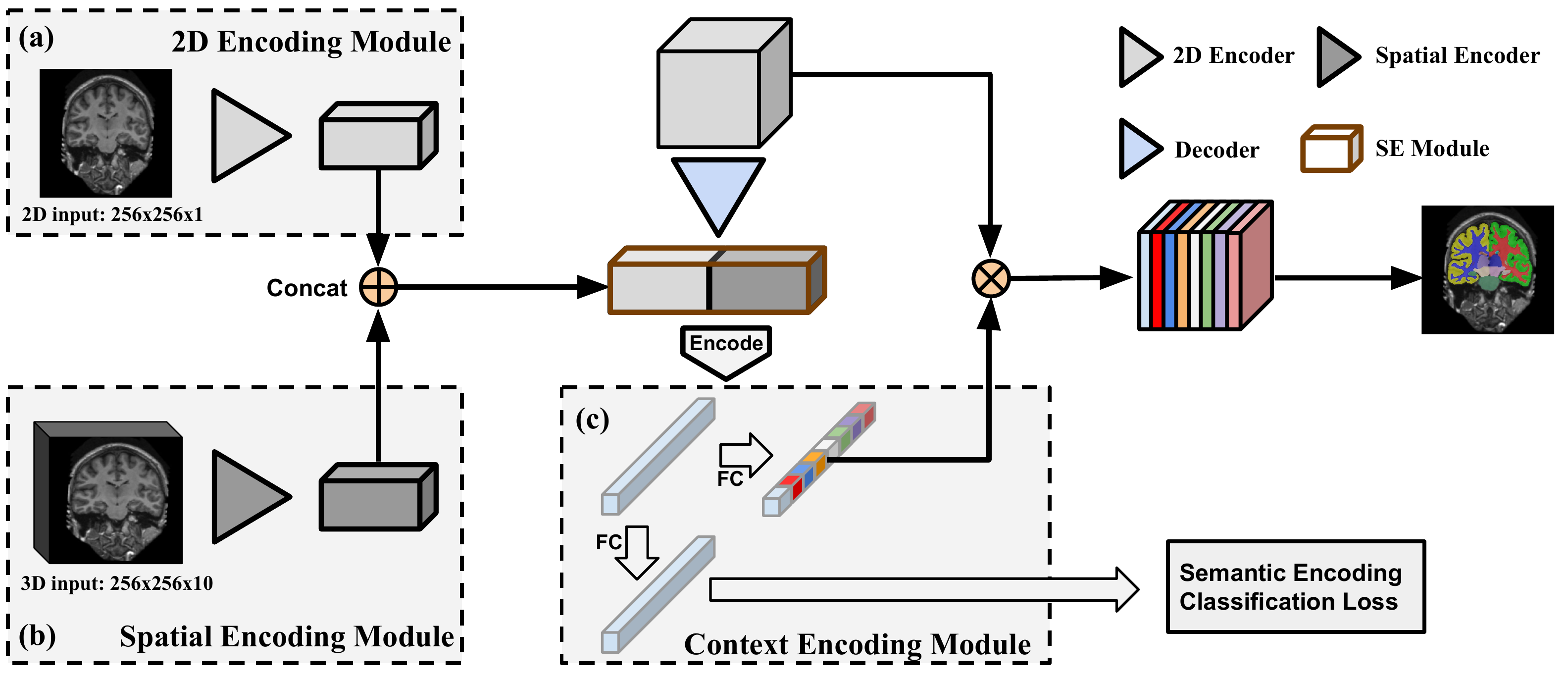}
\end{center}
  \caption{Overview of the proposed network structure. Given inputs of 2D image slice (a) and 3D image volume (b), we first pass each input to a densely-connected encoder to obtain both 2D and 3D intrinsic features. We then build a Context Encoding Module (c) on top of the fused features. The encode module contains an Encoding Layer to capture the encoded global semantic context and predict the scaling factor that highlight the class-dependent variation to the encoded semantics. In order to fully utilize the rich features extracted from Context Encoding Module, we employ Semantic Encoding Classification Loss to regularize the training. The final per-pixel prediction is obtained by a channel-wise multiplication from the scaling factor and the decoding features.}
\label{backbone}
\end{figure*}
%-------------------------------------------------------------------------

%------------------------------------------------------------------------
\subsection{Context Encoding}
\vspace{-0.1cm}
Context encoding module \cite{zhang2018context} was designed to capture the feature statistics as a global semantic context tuned by a semantic encoding loss. The semantic context serves as a scaling factor to selectively highlight or deactivate the class-dependent feature maps to facilitate the semantic segmentation. The context encoding strategy can be adapted to slice based brain structure segmentation naturally. By estimating if a brain structure appears in the slice to be segmented, it provides a global anatomical prior for the segmentation to squeeze the intensity ambiguity between structures with similar appearances, especially for fine-grained segmentation with small structures. 

Different from conventional context encoding strategy, in this study, our work focuses on both global semantic context and local dense features. We insert a modified context encoding module into the bottleneck part of the framework, and use the semantic contextual output of the module to recalibrate the features obtained by the decoder. Moreover, we combine 3D intrinsic spatial information with the global context encoding to pursue accurate segmentation.

\subsection{Squeeze \& Excitation block}

The idea of recalibrating the feature maps by context encoding information is inspired by the Squeeze \& Excitation (SE) networks for image classification \cite{hu2018squeeze}. The spatial and channel SE (sc-SE) blocks have also been recently equipped with FCN framework for brain structure segmentation in recalibrating QuickNAT \cite{roy2019recalibrating}. In existing studies, the SE blocks are usually placed after several convolution blocks to rescale the output feature map of the previous layer. While it serves as a self-attention to highlight discriminative feature maps for segmentation, it does not encode the global anatomical prior information explicitly. Different from the configurations of recalibrating QuickNAT, in our model, the global context encoding information is squeeze-and-excited from the bottleneck block, which interacts between encoder and decoder, to emphasize the global anatomical prior for the segmentation.

%-------------------------------------------------------------------------
\section{Feature-Fused Context-Encoding Net}

In this section, we present our proposed Feature-Fused Context-Encoding architecture, which comprises three encoding modules, as shown in Fig. \ref{backbone}. The first part is a 2D encoding module for extraction of feature information from the coronal plane. The second part is a spatial encoding module that extracts the intrinsic context features. The third part is a context encoding module that captures the global semantic contexts. Our network utilizes densely connected blocks inside the encoder and decoder structure, and takes sc-SE structure as self-attention to highlight feature maps.

\subsection{Feature Fusion Module}
%2D and 3D fused feature
Our proposed feature fusion module is composed of two parts: 2D encoding module in Fig. \ref{backbone}-a and 3D encoding module in Fig. \ref{backbone}-b. The 2D encoding module consists of a set of densely connected blocks, and each followed by a max-pooling block. By taking a single image slice as input, the 2D encoding module effectively extracts dense regional information from the coronal plane. Note that in such a 2D encoding setting, the goal of this module is to effectively generate feature semantics based on the image textures and intensities.

Context features obtained so far are not sufficient for a fine-grained structure segmentation, especially when the label has more than 100 classes. Therefore we utilize the auxiliary 3D spatial information to provide a more comprehensive measurement regarding neuroanatomy, specifically for small size structures such as angular and temporal gyrus \etc. Different from conventional 3D network, our proposed spatial encoding module takes the consecutive image slices as input. In this scenario, the third dimension of the input 3D images can be regarded as the $H\times W\times C$ stacked  2D image slices along the channel, rather than an $H\times W\times D\times C$ depth as in 3D volumes, where $H$ is the height, $W$ is the width, $D$ is the depth, and $C$ is number of channels. By regarding depth as channel from the stacked 2D image slices ($H\times W\times D$) shown in Fig. \ref{backbone}-b, the input for spatial encoding module is constructed in the same dimensionality as 2D input in Fig. \ref{backbone}-a. This strategy acquires intrinsic spatial context information with less computation compared with a 3D convolution module. This spatial encoding module follows the same structure as the encoder in Fig. \ref{backbone}-a.

%------------------------------------------------------------------------

\subsection{Context Encoding Module}
The context encoding module is designed in the bottleneck part of the frame work in Fig. \ref{backbone}-c. This module consists of an encoding layer, a fully connected layer and an activation function. The encoding layer is incorporated with the anatomical prior to capture the global semantic context. The global semantics obtained from encoding layer is passed through a fully connected layer followed by a sigmoid activation function. The scaling factor $\gamma$ of the class-dependent featuremaps are predicted from sigmoid function $\sigma(\cdot)$, \ie $\gamma = \sigma(We)$, where $W$ is the layer weights and $e$ is the encoding output. The network output is calculated as $Y = X \otimes \gamma$, where $\otimes$ is the channel-wise multiplication.

%------------------------------------------------------------------------

\subsection{Loss Function}
We utilize three loss functions during network training process: (i) a pixel-wise cross-entropy loss $L_{ce}$, (ii) a multi-class Dice loss $L_{dice}$ and (iii) semantic encoding classification loss $L_{sec}$. The pixel-wise cross-entropy loss provides a similarity estimation between output labels and manual labeled ground truth \cite{shore1980axiomatic}. Denote $p_{l}$ as the estimated probability of pixel $x$ belonging to class $l$, and $g_{l}$ as ground truth labels, the pixel-wise cross-entropy loss is
\begin{equation}
    L_{ce} = - \sum_{x} \omega(x)g_{l}(x)\log(p_{l}(x))
\end{equation}
The Dice loss proposed by Zou \etal \cite{zou2004statistical} measures the reproduciblity of the model by performing a pair-wise comparison between the generated label and ground truth. We adopt the the multi-class Dice loss setting from QuickNAT \cite{roy2019quicknat}, which is formulated as

\begin{equation}
    L_{dice} = - \frac{2 \sum_{x} p_{l}(x) g_{l}(x)} {\sum_{x}p_{l}^{2}(x) +\sum_{x}g_{l}^{2}(x)}
\end{equation}

In addition to standard training per-pixel loss that often concentrates on each individual pixel, while the global information represented by each class label are often overlooked. We therefore propose to utilize a Semantic Encoding Classification Loss (SEC-loss) to force network to focus on finding the corresponding global semantics from the given class label. This loss is applied on the global automatic semantics from the context encoding module as shown in Fig. \ref{backbone}-c. Such SEC-loss is learned as a binary cross entropy loss.

%-------------------------------------------------------------------------

\section{Experimental Results}

Recent research works generally use 27 coarse structure for MRI brain segmentation tasks \cite{roy2019quicknat,roy2017error,ronneberger2015u,roy2019recalibrating}. The differences between the coarse and the fine-grained labels are shown in Fig. \ref{view}. Our network is desiged for a fine-grained MRI brain structure segmentation on 134 labels. We compare our model with the baseline network \cite{roy2019quicknat}, and validate our network on coarse structure segmentation with 27 labels so as to have a comprehensive comparison. Our network attains evident improvement over the state-of-the-art approaches \cite{roy2019quicknat,roy2017error,ronneberger2015u}.

%-------------------------------------------------------------------------
%table for result section
\begin{center}
    \renewcommand{\arraystretch}{1}
    \begin{table*}[ht]
      \centering
      \begin{tabu}{c c c c c c | c}
      \hline\tabucline[2pt]{-}
    \multicolumn{2}{c}{Inputs}  & \multicolumn{2}{c}{Decoder Backbone}  & \multirow{2}{*}{Class Weight} & \multirow{2}{*}{Batch Size} & \multirow{2}{*}{Dice Score} \\
    2D & 3D & number blocks & block type &  &  & \\ \tabucline[2pt]{-}
      \hline
      \checkmark 	&           & 2 	& Conv (Unet-like)	&    & 20 &  0.768\\
      \checkmark 	&           &2 		& Dense	&           & 15 &  0.834\\
      \checkmark 	&           &4 		& Dense	&\checkmark & 8  &  0.870\\
      \checkmark 	&           &4 		& Dense	&           & 8  &  0.884\\
      \checkmark 	&\checkmark &4 		& Dense	&\checkmark & 6  &  0.867\\
      \checkmark 	&\checkmark &4 		& Dense	&           & 6  &  \textbf{0.888}\\
      \hline \hline
                    & QuickNAT  &4 		& Dense	&\checkmark & 10 & 0.851 \\
                    & QuickNAT  &4 		& Dense	&           & 10 & 0.876 \\
        \hline \tabucline[1pt]{-}
      \end{tabu}
    \caption{Inference strategy on 2012 MALC test set with coarse (27) segmentation labels for different network settings. Number blocks represents the number of decoder blocks consists with a up-sampling layer. The input contains the 2D slices for 2D encoding module and 3D volumes for spatial encoding module. \checkmark indicates the presence of the subject. The last two rows shows the implemented baseline network with a QuickNAT backbone.}
    \label{diffconv}
    \end{table*}
\end{center}
% %-------------------------------------------------------------------------
%-------------------------------------------------------------------------
% table for result section
\begin{center}
    \renewcommand{\arraystretch}{1}
    \begin{table*}[h]
      \centering
      \begin{tabu}{c c c c| c}
      \hline\tabucline[2pt]{-}
        Methods  & Processing time  & Subject Training  & Subject Testing  & Dice Score (\textbf{coarse}) \\ \tabucline[2pt]{-}
      DeepNAT \cite{wachinger2018deepnat}  & 1 hrs  & 20 & 10  & 0.891 (25 structures) \\
      SD-Net  \cite{roy2017error}          & 7 secs & 15 & 10  &  0.850 \\
      U-Net \cite{ronneberger2015u}        & 7 secs & 15 & 10  &  0.810 \\
      QuickNAT \cite{roy2019quicknat}      & 7 secs  & 15 & 15   & 0.876 \\
      \hline
      Ours (2D)    & 6 secs   & 15 & 15  & 0.884 \\
      Ours (Fused)    & 6 secs   & 15 & 15   & 0.888 \\
       
      \hline \tabucline[1pt]{-}
      \end{tabu}
    \caption{Comparison of Dice scores on 2012 MALC test set with coarse (27) segmentation labels. The second column shows the approximate time to generate semantic labels for the whole input volume. Third and fourth column show the number of subjects used for each network. The second last row represents 2D encoding module only, and the last row represents 2D encoding module and a spatial encoding module.}
    \label{dice_co}
    \end{table*}
\end{center}
%-------------------------------------------------------------------------

\vspace{-1.2cm}
\subsection{Dataset}

We evaluate our network on 2012 Multi-Atlas Labelling Challenge (MALC) dataset. The MALC datast contains 134 fine-grained structure segmentation labels. Segmenting those 134 structures is a challenging task due to the similar appearance of each structure. The MALC dataset is part of OASIS data \cite{marcus2007open} and contains MRI T1 scans from 30 subjects with 134 manual segmentation labels \cite{landman2012miccai} provided by Neuromorphometrics, Inc\footnote{http://www.neuromorphometrics.com/}. In this challenge, 15 subjects are selected for training and 15 subjects are used for testing. Both training and testing lists are provided by \cite{landman2012miccai}. Note that we did not apply any preprocessing schemes such as skull-stripping, intensity re-normalization \etc. 
All input scans are resampled into an isotropic volume of 1mm$^{3}$ by FreeSurfer \cite{fischl2012freesurfer}.

\subsection{Implementation Details}
Our network takes 2D image slices of $256 \times 256$ and a 3D image volume of $256 \times 256 \times 10$ as inputs. Both inputs are selected in a coronal view. We employed the learning rate scheduling ``poly" $lr = baselr * (1 - \frac{iter}{iter_{total}})^{p}$ from \cite{chen2018deeplab}, with a start learning rate of 0.01 for coarse structures and 0.02 for fine-grained structures. The weight decay rate is set to $1\times 10^{-4}$ for all tested models. A dropout \cite{srivastava2014dropout} rate of 0.1 is applied to each densely connected block. We used a weight factor of 1 for dice loss and 0.1 for SEC-loss. In section \ref{design_choice}, we studied the optimal class weights for pixel-wise cross-entropy loss. The model is trained for 100 epochs in total. All experiments are performed on a single NVIDIA TITAN XP GPU with 12 GB of RAM.

%-------------------------------------------------------------------------
%table for result section
\begin{center}
    \renewcommand{\arraystretch}{1.2}
    \begin{table}[h]
      \centering
      \begin{tabu}{c | c}
       \hline\tabucline[2pt]{-}
    Methods  & Dice Score (\textbf{fine-grained}) \\
    \tabucline[2pt]{-}
       \hline
       3D SegNet \cite{de2015deep}      & 0.725 \\
       QuickNAT \cite{roy2019quicknat} & 0.687 \\
       \hline
       Ours (2D) & 0.727 \\
       Ours (Fused) & \textbf{0.733} \\
       
       \hline \tabucline[1pt]{-}
      \end{tabu}
    \caption{Comparison of Dice scores on 2012 MALC test set with fine-grained (134) segmentation labels.}
    \label{dice_fine}
    \end{table}
\end{center}
%-------------------------------------------------------------------------
%-------------------------------------------------------------------------
\begin{figure*}[!h]
\begin{center}
\includegraphics[width=1\textwidth]{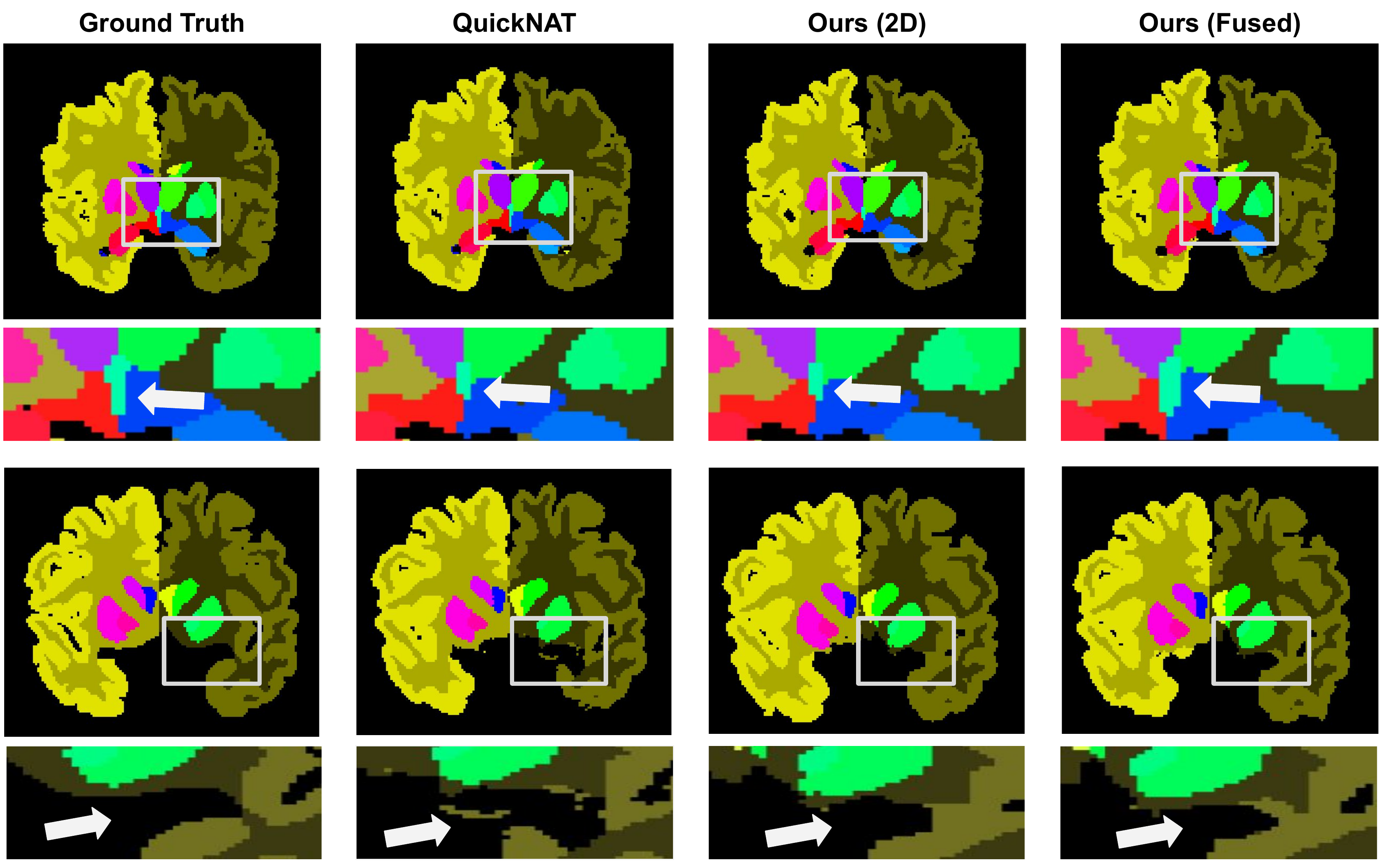}
\end{center}
  \caption{Visualization of coarse structure segmentation with 27 labels. We visualized our proposed network with two encoding modules. The results are compared with ground truth and QuickNAT in the pixel level. The second and fourth rows are the details of structure localized by bounding boxes.}
\label{comp_coarse}
\end{figure*}
%-------------------------------------------------------------------------
\vspace{-1cm}

\subsection{Network Backbone}
We employed 4 densely connected blocks for encoder structure, each block is followed by a max-pooling layer. The output obtained from 2D encoding module in Fig. \ref{backbone}-a and spatial encoding module in Fig.\ref{backbone}-b are concatenated together as the fused feature. The fused feature is simultaneously utilized as the input for a context encoding module Fig. \ref{backbone}-c and a decoder. The skip connection layers are used between 2D encoding module and the decoder. We applied the spatial and channel Squeeze-and-Excitation (sc-SE) \cite{roy2019recalibrating} for each encoding and decoding dense blocks, and used sc-SE structure for the fused feature to extract regional context from 2D inputs and spatial context from 3D inputs. Our network is expected to achieve especially better performance for small objects and on fine-grained structures.

\subsection{Experimental Setting} \label{design_choice}
In this section, we introduced our implemented baseline network. We further studied the network performance on various decoder structure and class weights to select the model backbone. We then compared the effectiveness of the feature fusion model and a single 2D encoding model in this coarse structure segmentation task, as shown in Table \ref{diffconv}. All models are evaluated on 2012 MALC dataset with 27 coarse labels. The comparison with state-of-the-art networks are shown in Table \ref{dice_co}.

\textbf{Baseline:}
We adopted the backbone structure of QuickNAT \cite{roy2019quicknat} as the baseline. It is worth noting that, original QuickNAT \cite{roy2019quicknat} trained the network on multi-view 2D slices along 3 principle axes (\ie coronal, sagittal, axial). They then used additional 581 images as auxiliary labels for pre-training, and obtained final results after fine tuning. In contrast, we used a single coronal view image slice as input to train our baseline QuickNAT on manual labels without any pre-train dataset. Similar with the prior work \cite{roy2019recalibrating}, we added the spatial and channel Squeeze-and-Excitation module inside each densely connected block for baseline. We kept such training setting the same for the baseline and all the tested networks. Performance of the baseline is shown in the last two rows of Table \ref{diffconv}.

%-------------------------------------------------------------------------
\begin{figure*}[!h]
\begin{center}
\includegraphics[width=\textwidth]{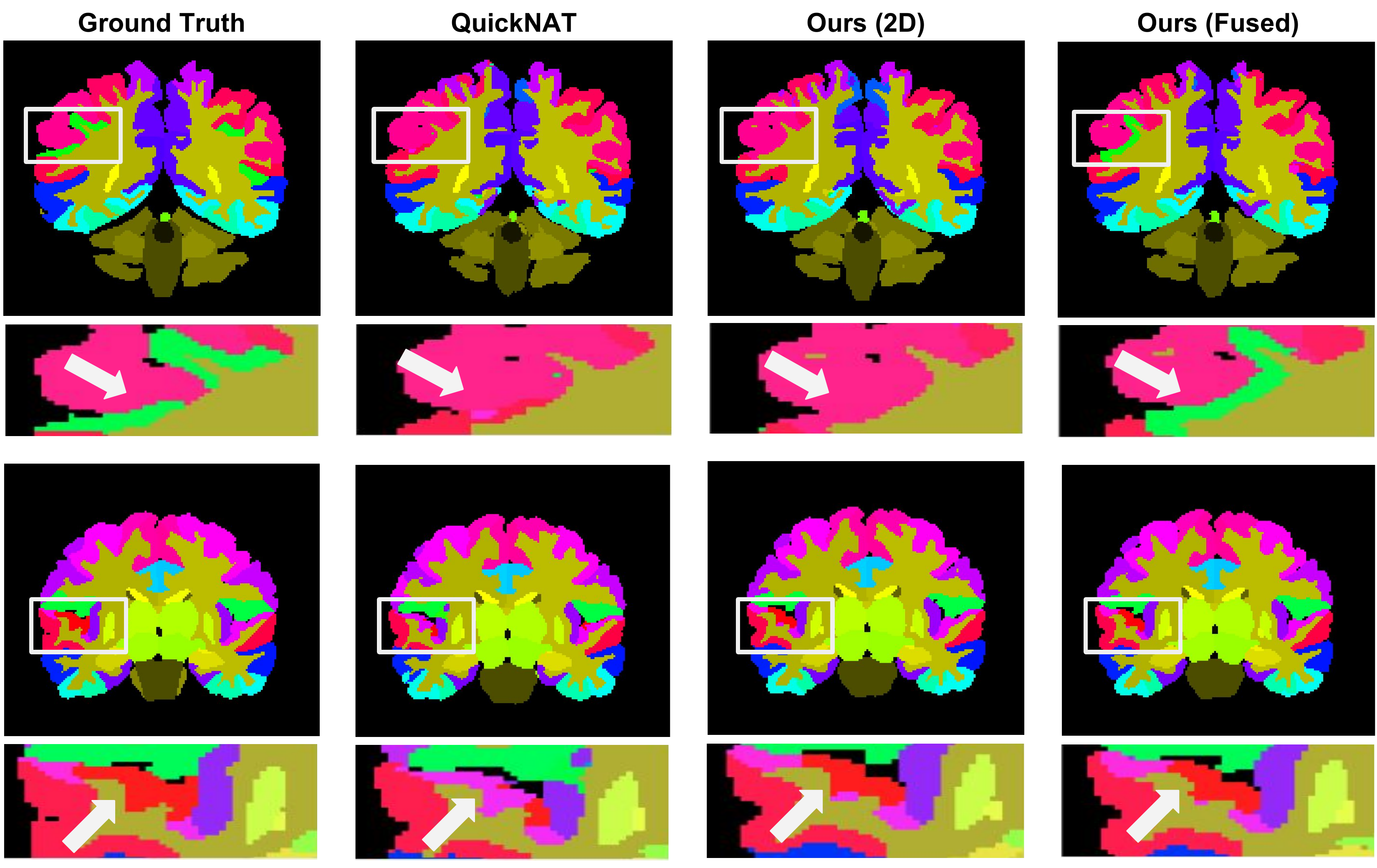}
\end{center}
  \caption{Visualization of fine-grained structure segmentation with 134 labels. We visualized our proposed network with two encoding modules. The results are compared with ground truth and QuickNAT in the pixel level. The second and fourth rows are the details of structure localized by bounding boxes.}
\label{comp_fine}
\end{figure*}
%-------------------------------------------------------------------------

\textbf{Decoder Structures:}
As shown in Fig. \ref{backbone}, the informative regional and spatial context semantics from the bottleneck layer are passed through the decoder module. An effective decoder structure is critical for the performance of network. In this section, we studied several decoder structures shown in the first four rows of Table \ref{diffconv}, and the best architecture is selected as the decoder backbone for the rest of experiments. The decoder structures are tested on 2D input image slice only for the coarse segmentation task.

Since the encoder in our proposed network contains 4 down-sampling layers, the decoder module should also consist 4 up-sampling layers so as to keep the same resolution. The network was first tested on the decoder with 2 layers of convolutional blocks followed by a bilinear up-sampling ratio of 4. We tested both Unet-like blocks \cite{ronneberger2015u} and Densely-connected blocks \cite{huang2017densely}, and found that the densely connected blocks is 6.6\% higher than Unet-like blocks in terms of the mean Dice Score.. As the densely extracted features decode a more precise contextual and spatial semantics, we propose a decoder module with 4 layers of densely connected blocks, which achieves the best mean Dice Score of 0.884 among tested structures, as shown in Table \ref{diffconv}.

\textbf{Class Balanced Weights:}
The class weights were introduced by the prior works \cite{roy2017error,roy2019quicknat} to compensate the relative contribution of per pixel segmentation. The frequently appeared classes are given by a relative small weights, whereas the less appeared classes are having a higher weights in pixel-wise cross-entropy loss calculation. In this section, we studied the effect of class balanced weights on two conditions of different inputs shown in the middle four rows of Table \ref{diffconv}. We implemented the proposed class weight on three network models: (i) 2D input only, (ii) 2D and 3D inputs (iii) QuickNAT backbone. For all the tested models, we observed an average reduction of 1.67\% in the loss function by adding the class balanced weights. One of the possible reasons is that most of the coarse structure labels in the coronal view are more evenly distributed, whereas the class balanced weights lead to better performance when the input image classes are extremely imbalanced. Furthermore, a slightly change in the loss setting could yield different results, which might be another reason for the decline in performance.

\textbf{Comparison with Encoding Modules:} \label{3Dmod}
In this section we tested our network on coarse structures with 27 segmentation labels. We investigated the effectiveness of both 2D encoding module and 3D encoding module shown in Table \ref{diffconv}. The goal of the spatial encoding module is to capture the spatial information from axial and sagittal plane along the coronal direction using the input 3D volumes. Specifically, we want to examine the effectiveness of the spatial encoding module in coarse label structures. All models are trained in the same settings for a fair comparison. We observe that for coarse structure segmentation, the fused model yields the better results with 0.4\% increment with respect the 2D encoding model, and a 1.2\% increment compared to the baseline. The fused encoding model shows a better performance on brain segmentation especially on detailed structures and sharp inter-class edges. 

The performance comparison of our network on 2012 MALC data for coarse structure segmentation is shown in Table \ref{dice_co}. Both of our proposed network in 2D and fused encoding module is able to achieve a higher mean Dice Score compare to the-state-of-the-art methods, with the less amount of time. This results further demonstrate that our implemented context encoding module could effectively capture the anatomical prior information. The comparison between our proposed models, baseline and ground truth are shown in Fig. \ref{comp_coarse}. The bounding boxes show the third ventricle (\ie fluid-filled cavities). The highlighted regions show the segmented labels produced by the models. The proposed model with a fused (2D and spatial) encoding module generates more closed segmentation for the rear part of the third ventricle (small region), and produce the better details near the sharp edges of the brain. Moreover, both of our proposed network produce a more accurate segmentation labels compare to our baseline \cite{roy2019quicknat}.

\subsection{Evaluation on Fine-Grained Structures} \label{eval}

Different from the coarse structures, the fine-grained structures consists a more complicated shapes and textures as shown in Fig. \ref{view}, from which the complex region details and sharp edges among inter-class labels are ignored by coarse structure segmentation. The fine detailed structures such as pallidum (structure within the basal ganglia), frontal operculum (part of the the frontal lobe that overlies the rostrodorsal portion of the insula), lingual gyrus (structure linked to processing vision) \etc contain the labels in a smaller size. The symmetrical structures are assigned with different labels on left and right side of the lobe. Specifically, we want to examine the effectiveness of the spatial encoding module in fine-grained label structures. With the structural distinction between coarse and fine-grained labels, this tasks would exhibit different performance compare to Section \ref{3Dmod}.

We compared the proposed network with the-state-of-art approaches \cite{roy2019quicknat,de2015deep} and evaluated on mean Dice Score, as shown in Table \ref{dice_fine}. For fine-grained structure segmentation, the fused encoding module shows a 0.6\% improvement over 2D encoding module and with mean Dice Score of 73.3\%, which proves that the spatial encoding module achieves more precise segmentation on the complicated structure. The perceptual results of fine-grained structure segmentation are shown in Fig. \ref{comp_fine}. The difference between our proposed models and the baseline can be observed in the bounding box regions. The left angular gyrus labeled with green are only captured by our fused encoding module, while the left transverse temporal gyrus labeled with red are captured by both of our proposed models. Based on such observations, we can conclude that our 2D and spatial model effectively captures more fine-detailed structures compared with 2D models and the baseline.

%-------------------------------------------------------------------------
\section{Discussion}
Our method is able to produce an accurate MRI Brain segmentation for fine-grained structure. We used MultiAtlas Labelling Challenge (MALC) dataset to train and evaluate our models. We tested our network decoder module with various backbone structures and found a 4 layers of densely connected blocks yield the best performance. We then studied the impact of class weights, and found that our network achieves a bigger performance without class weights. We performed comparative experiments with the state-of-the-art methods and showed the prominent improvement of our model on both coarse and fine-grained structure segmentation. Our network is able to achieve a higher Dice Score within an efficient processing time. The proposed fused encoding module are essential for generating the fine-detailed labels on the structure with complex region details and sharp edges.

Current proposed method only utilizes the spatial features obtained along the coronal direction, one of the future improvement for this work is to implement a multi-view spatial encoding module to further facilitate brain structure segmentation on fine-grained labels. Another important issue is overfitting caused by densely connected blocks. How to prevent network overfitting remains a future work.

%-------------------------------------------------------------------------
\section{Conclusion}
In this paper, we introduced an end-to-end Feature-Fused Context-Encoding model for MRI Brain segmentation. Our model learns semantic contexts through a global anatomical prior and utilizes the 3D spatial information to achieve the fast and accurate volumetric segmentation. We transformed our feature fusion module into a 2D encoding module and a spatial encoding module to extract both regional and spatial features. We proposed to use a semantic encoding classification loss to enforce the learning of global information from the input slices. We tested our model on 2012 Multi-Atlas Labelling Challenge dataset, and achieved prominent improvement compare to the state-of-the-art approaches on both coarse structure segmentation and fine-grained structure segmentation.

%-------------------------------------------------------------------------

{\small
\bibliographystyle{ieee}
\bibliography{egbib}
}

\end{document}